\documentclass[10pt,twocolumn,letterpaper]{article}
\usepackage[pagenumbers]{cvpr}
\usepackage[T1]{fontenc}
\usepackage{amsmath,amssymb,booktabs,graphicx,multirow}
\definecolor{cvprblue}{rgb}{0.21,0.49,0.74}
\usepackage[pagebackref,breaklinks,colorlinks,allcolors=cvprblue]{hyperref}
\def\confName{CVPR}\def\confYear{2027}
\newcommand{\figone}[4]{\begin{figure*}[t]\centering\includegraphics[width=\textwidth]{assets/#1.pdf}\caption{#2}\label{#3}\end{figure*}}
\title{Zero-Shot Compositional Generalization for Embodied Manipulation}
\author{%
\begin{minipage}{0.96\textwidth}
\centering
\fontsize{11}{13}\selectfont
Baicheng Li\textsuperscript{1,*}\quad Wenkang Qin\textsuperscript{2,*}\quad Youqi Pan\textsuperscript{1}\quad Dongxiao Mao\textsuperscript{2}\\[2pt]
Yukun Zhou\textsuperscript{2}\quad Chuanfu Shen\textsuperscript{3}\quad Dong Wu\textsuperscript{1}\quad Shunkai Zhou\textsuperscript{1}\\[2pt]
Zike Yan\textsuperscript{4}\quad Wei Sui\textsuperscript{2,*}\quad Hongbin Zha\textsuperscript{1,*}\\[5pt]
{\small
\textsuperscript{1}Peking University\quad\textsuperscript{2}D-Robotics\\[2pt]
\textsuperscript{3}Shenzhen Institute for Advanced Study,\\
University of Electronic Science and Technology of China\\[2pt]
\textsuperscript{4}Hong Kong Embodied AI Lab,\\
The Chinese University of Hong Kong (CUHK)
}
\end{minipage}
}

\hypersetup{pdftitle={Zero-Shot Compositional Generalization for Embodied Manipulation},pdfauthor={Baicheng Li, Wenkang Qin, Youqi Pan, Dongxiao Mao, Yukun Zhou, Chuanfu Shen, Dong Wu, Shunkai Zhou, Zike Yan, Wei Sui, Hongbin Zha}}

\makeatletter
\g@addto@macro\@maketitle{
  \begin{minipage}{\textwidth}
    \centering
    \includegraphics[width=0.60\textwidth]{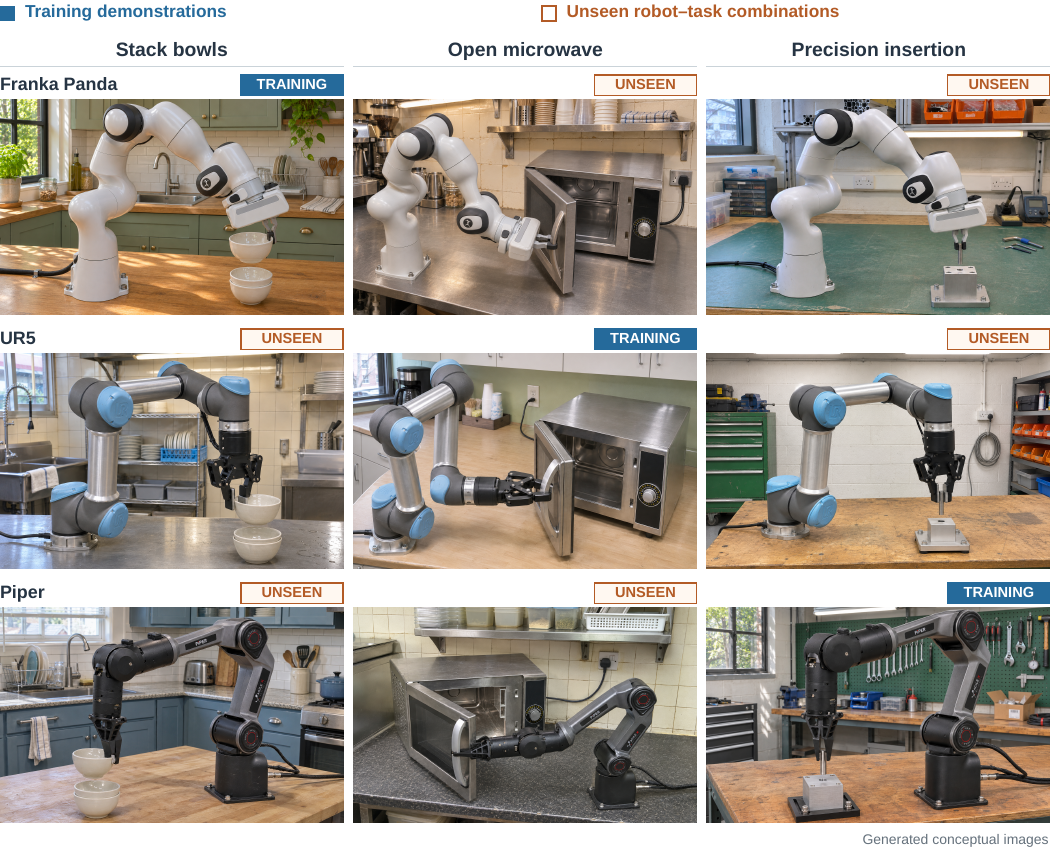}
    \captionof{figure}{Known robots and skills, unseen combinations. Three illustrated training pairings (blue labels) leave six unseen robot--task combinations (orange outlined labels). Rows show Franka Panda, UR5 and Piper. Columns show three skills. Generated conceptual images illustrate the setting.}\label{fig:teaser}
  \end{minipage}
}
\makeatother
\begin{document}\maketitle
\begin{figure*}[t]\centering
\includegraphics[width=\textwidth]{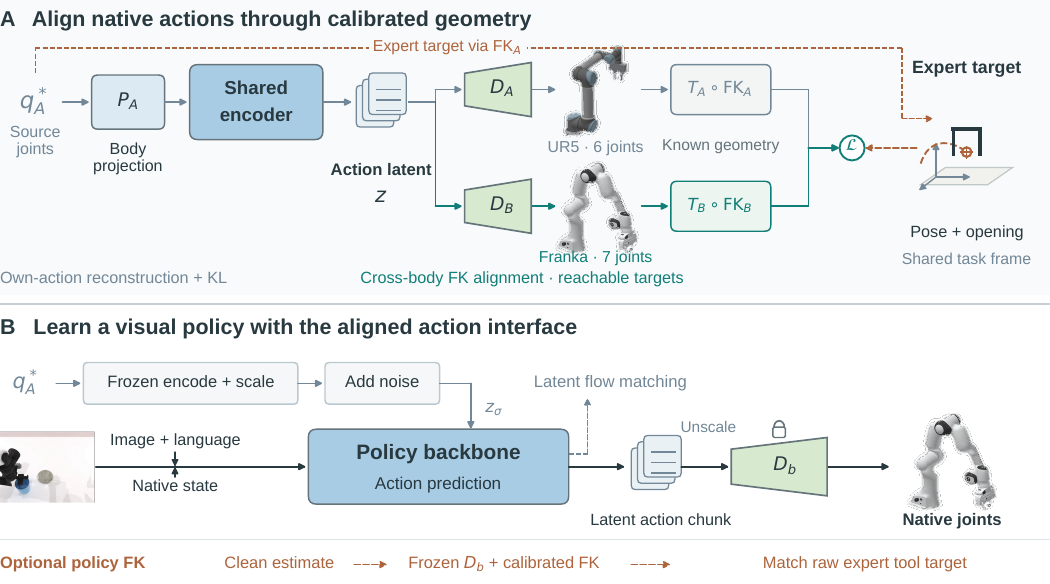}
\caption{Shared tool-space supervision. (A) A shared encoder connects native action decoders. Source commands define a calibrated expert tool target that supervises eligible foreign realizations. (B) A visual policy learns shared actions from observed pairings and executes through the target robot's frozen native decoder. The lower dashed branch adds optional policy-stage FK supervision. Robot poses and the tool target are schematic.}\label{fig:framework}
\end{figure*}

\begin{abstract}
Robot demonstrations are collected on different embodiments, leaving many robot--skill combinations unobserved. Pooling these data shares visual knowledge, while joint labels remain tied to the demonstrating kinematic chain. We propose a geometry-supervised training framework for zero-shot composition of familiar robots and skills. The central insight is that a demonstrated command provides a tool-space target for every robot that can realize it. Calibrated forward kinematics supplies this supervision to foreign native action interfaces without their joint labels for that skill. Native reconstruction retains demonstrated configurations, and positive reachability checks select targets for cross-robot supervision. A visual policy learns to predict shared actions and executes through the target robot's native decoder. During policy learning, one prediction receives geometric supervision through both source and foreign native realizations. Matched supervision controls, repeated policy training and evaluations across three backbones and two simulators demonstrate improved transfer to missing robot--task combinations.
\end{abstract}

\section{Introduction}
Suppose one robot learns to place a cup on a shelf and another learns to place containers in a drawer. Each has practiced grasping, transporting and releasing objects. A shared policy has also seen both tasks. Yet asking the second robot to place the cup introduces a missing combination of a familiar robot and a familiar skill. This gap is common when robot datasets are assembled from separate platforms, environments and collection efforts. Collecting every task on every robot scales with the product of the two sets. Reusing demonstrations requires a way for supervision from an observed combination to inform a missing one.

The action labels make this difficult. A joint vector describes how the demonstrating robot moves. Its entries depend on joint count, link geometry, joint conventions and installation. Even when two arms have the same number of joints, equal numerical commands can move their tools differently. A shared model can learn useful transfer from images, language and repeated task structure, but native labels provide no explicit correspondence between these different motions. Padding labels to a common width makes them compatible with a network input or output. The remaining learning problem is to connect their physical meanings.

Known robot geometry supplies a physical correspondence. A demonstrated joint command determines the calibrated tool's position, orientation and opening in a shared task frame. For another robot, this target constrains the effect of its action while leaving its joint solution to its own kinematic chain. One source demonstration can therefore supervise several native action interfaces, including those without demonstrations of the source task. The required information is the robots' geometry and calibration, together with evidence that a foreign chain can reach the target. Figure~\ref{fig:teaser} illustrates how this supervision can connect familiar robots and skills across missing combinations.

Our approach learns native action interfaces around this supervision path. A source action is encoded into shared coordinates and decoded by both its own interface and a foreign robot's interface. Native reconstruction anchors each robot to its demonstrated commands. Cross-robot geometric supervision trains the foreign decoder to realize the source tool target, using a positive reachability check to select admissible targets. A visual--language policy subsequently learns the shared action coordinates and executes through the appropriate native decoder. Geometry establishes a relation among action realizations. The policy must still learn from observations and language when to approach, grasp, transport and release, and closed-loop execution determines whether those decisions complete the task.

Supervision connects missing robot--task edges through action realization. A source demonstration supplies the skill's tool targets. A foreign robot supplies its kinematics and native demonstrations of other skills. Together, they train its interface to realize the source tool targets without a paired joint trajectory. Native reconstruction alone leaves this cross-interface relation unspecified. Matched-exposure controls test whether learning the relation improves closed-loop execution on missing combinations.

Our contributions are twofold. First, we learn heterogeneous whole-arm action interfaces from demonstration-derived tool targets. A source command supervises a foreign native decoder on reachable targets, while native reconstruction anchors demonstrated configurations. An interface-level ambiguity analysis explains why native reconstruction alone leaves foreign action semantics underspecified. Second, we supervise policy predictions through source and eligible foreign realizations using the same demonstration-derived tool targets. Matched-exposure controls, action-level diagnostics and repeated policy learning test these two supervision paths across three backbones and two simulators on missing robot--task combinations.

\section{Related Work}
\paragraph{Sharing demonstrations across robots.}
Open X-Embodiment aggregates heterogeneous robot experience, and Octo uses flexible tokenization and diffusion readouts to learn from multiple datasets~\cite{embodimentcollaboration2025openxembodimentroboticlearning,octomodelteam2024octoopensourcegeneralistrobot}. HPT combines dataset-specific stems and heads with a shared trunk~\cite{wang2024scalingproprioceptivevisuallearningheterogeneous}. X-VLA uses data-source prompts and an aligned end-effector representation~\cite{zheng2025xvlasoftpromptedtransformerscalable}. These designs provide ways to pool experience and share model capacity. Their action interfaces also determine which correspondences are supplied directly to learning. RT-X retains dataset-dependent control coordinates, while UMI expresses trajectories relative to the current tool~\cite{embodimentcollaboration2025openxembodimentroboticlearning,chi2024universalmanipulationinterfaceinthewild}. Our study focuses on the supervision supplied by a geometric correspondence when expert coverage omits robot--task combinations.

\makeatletter
\raggedbottom
\AddToHookNext{shipout/after}{\global\let\@textbottom\relax}
\makeatother
\paragraph{Geometric and learned action interfaces.}
UniVLA learns task-centric discrete actions from video, while Latent Action Diffusion aligns hand and gripper representations using retargeted pairs and contrastive learning~\cite{bu2025univlalearningacttaskcentric,bauer2026latentactiondiffusioncrossembodiment}. XL-VLA aligns hand interfaces with differentiable FK and evaluates missing hand--task combinations~\cite{jiang2026crosshandlatentrepresentationvisionlanguageaction}. It also reports xArm/G1 co-training and releases arm-realization utilities, while its autoencoder learns hand latents and passes seven arm coordinates through~\cite{xlvlareleasedcode}. Our native decoders learn heterogeneous whole-arm realizations, using calibrated tool targets to supervise different kinematic chains. KITE learns shared contact-intent representations and native kinematic decoders from synthetic geometry, followed by policy learning through frozen interfaces~\cite{wang2026kitedecouplingkinematicsinteraction}. Action-prior learning pretrains motion representations before visual-language alignment~\cite{jing2026learningactionpriorscrossembodiment}. We derive foreign whole-arm supervision from demonstrated tool targets and test its effect on missing robot--task edges under matched exposure.

\paragraph{Learning visual control through action interfaces.}
ACT predicts joint-action chunks using a conditional VAE, Diffusion Policy denoises action sequences, and OpenVLA predicts discretized control coordinates~\cite{zhao2023learningfinegrainedbimanualmanipulation,chi2024diffusionpolicyvisuomotorpolicy,kim2024openvlaopensourcevisionlanguageactionmodel}. The $\pi_0$ family uses flow-based action experts, with $\pi_{0.5}$ accommodating joint and end-effector supervision~\cite{black2026pi0visionlanguageactionflowmodel,intelligence2025pi05visionlanguageactionmodelopenworld}. OpenWAM couples video and action generation, while Fast-WAM retains video co-training and avoids future-video denoising during action inference~\cite{wang2026openwamopenmodularexploration,yuan2026fastwam}. These policy families provide platforms on which to test the action-supervision question. GRASP and MorphIK further establish FK losses on clean action estimates in flow-based learning~\cite{ren2026grasp,clasmeier2026morphik}. We evaluate this additional gradient path through frozen multi-robot interfaces separately from the geometric supervision used to learn those interfaces.

\section{Problem Formulation}
\subsection{Missing Combinations of Known Robots and Tasks}
Let $\mathcal B$ be a set of robots and $\mathcal K$ a set of tasks. Demonstrations define edges in a bipartite robot--task graph. Action-interface pretraining observes edges $\mathcal O_{\mathrm{codec}}$, and visual policy learning observes $\mathcal O_{\mathrm{policy}}$. We evaluate held-out combinations $\mathcal H$ satisfying
\begin{equation}
 \mathcal H\cap(\mathcal O_{\mathrm{codec}}\cup\mathcal O_{\mathrm{policy}})=\emptyset.
 \label{eq:heldout}
\end{equation}
Every tested robot and task individually appears in policy training. Each robot therefore has native demonstrations, and each tested skill has been demonstrated by another robot. Zero-shot transfer means executing their missing combination without its expert demonstrations in either training stage. Evaluation episodes are also held out. Kinematic descriptions, joint limits, installation transforms and tool calibration are known for every robot.

A source demonstration provides images $o$, language $\ell$, current native state $q_s$, and future native commands $q^*_{s,1:H}$. The policy maps current observations and a task instruction to native commands for the executing robot. Cross-robot supervision may use tool targets derived from the source demonstrations. It supplies no target-robot images or expert joint trajectory for the missing combination. This distinction defines the permitted information transfer across the graph. Source-derived geometric targets can constrain target action realization while visual task behavior is learned from the observed edges.

\subsection{A Shared Target for Different Joint Realizations}
For arm $a$ of robot $b$, define its calibrated tool pose in task frame $\mathcal W$ as
\begin{equation}
 T_{b,a}(q)={}^{\mathcal W}T_{b,a}^{\mathrm{root}}\,F_{b,a}(q)\,T_{b,a}^{\mathrm{tool}}.
 \label{eq:fk}
\end{equation}
The three factors specify installation, forward kinematics and the flange-to-tool calibration. A source command yields $y_{s,a}=(p_{s,a},R_{s,a},g_{s,a})$, where $g$ is gripper opening normalized to $[0,1]$. Corresponding left and right arms share targets. The same $y_{s,a}$ can supervise robots with different joint dimensions because their predictions are compared after applying their own calibrated kinematics.

Whole-arm realization introduces two constraints. First, reachable tool poses depend on the target chain, installation and joint limits. We supervise a foreign arm only when a numerical IK procedure finds a finite, joint-limited solution within the specified pose tolerances. This positive witness supports the particular target, while an unsuccessful search leaves it unsupervised. Second, redundant arms can realize one pose with multiple joint configurations. Each robot's reconstruction supervision anchors its decoder to demonstrated native behavior, and foreign geometry constrains the tool effect of decoding a source action. The resulting target is a pose-level constraint. Trajectory continuity, collision avoidance and successful contact remain properties to assess in execution.

\section{Transferring Supervision Through Robot Geometry}
\subsection{From a Demonstrated Action to a Foreign Interface}
Figure~\ref{fig:framework} shows the supervision path. Here $q_s$ denotes a demonstrated action command. An encoder $E_s$ maps it into shared action coordinates $z$. A robot-specific decoder $D_b$ produces native commands with robot $b$'s joint dimension and limits. The central training relation is
\begin{equation}
 q_s\xrightarrow{E_s}z\xrightarrow{D_b}\widehat q_b,
 \qquad T_{b,a}(\widehat q_b)\approx T_{s,a}(q_s).
 \label{eq:transfer}
\end{equation}
The right-hand target comes from the source command. Each foreign decoder produces its own native command, and its calibrated FK maps that prediction into the same task frame. The resulting pose error can train the decoder even when no foreign joint label exists. Thus the supervision crosses robots through a shared physical effect. Native reconstruction supplies the demonstrated configurations from which each decoder learns its realization prior.

Our implementation uses body-specific input projections, a shared variational encoder and native decoders. Each action instant receives a 32-dimensional representation. Chunks are sequences of these representations. For a source command, a sample $z\sim q_\phi(z\mid q_s,s)$ is decoded through its native interface and selected foreign interfaces. The source terms combine standardized native reconstruction, source tool-pose reconstruction and KL regularization,
\begin{equation}
 \mathcal L_{\mathrm{own}}=\mathcal L_q+\eta\mathcal L_{\mathrm{src}}+
 \beta\mathcal L_{\mathrm{KL}}.
\end{equation}
For sampled source instants $\mathcal I_s$ and a selected foreign robot $b$, cross-robot supervision is
\begin{equation}
 \mathcal L_{\mathrm{foreign}}=
 \frac{1}{2|\mathcal I_s|}\sum_{i\in\mathcal I_s}\sum_{a=1}^{2}
 m_{i,b,a}\,d\bigl(y_{b,a}(D_b(z_i)),y_{s,a}(q_{s,i})\bigr)
 \label{eq:foreign-codec}
\end{equation}
Here $m$ denotes the positive reachability witness, and $d$ sums scaled squared position, chordal rotation and normalized-opening errors. The denominator counts attempted arm targets. Training minimizes $\mathcal L_{\mathrm{own}}+\gamma\mathcal L_{\mathrm{foreign}}$. Hyperparameters, sampling and witness-solver settings appear in the supplementary implementation details.

At the interface-function level, consider a robot-dependent signed permutation $P_b$ of the action coordinates. Transform its posterior by $P_b$ and replace its decoder with $D'_b(z)=D_b(P_b^{\mathsf T}z)$. Native decoded-action distributions and KL to an isotropic Gaussian prior are unchanged. Foreign decoding becomes
\begin{equation}
 D'_b(P_sz_s)=D_b(P_b^{\mathsf T}P_sz_s),
 \label{eq:native-ambiguity}
\end{equation}
which can change the realized tool motion. Native reconstruction and prior regularization therefore leave the physical meaning of foreign decoding underspecified. Parameter sharing supplies an inductive bias, and cross-robot geometric supervision supplies an explicit target for its tool effect.

Native reconstruction encourages the interface to retain demonstrated distinctions among redundant configurations with the same tool pose. Foreign supervision constrains the decoded tool effect on reachable targets. We measure both foreign realization and downstream task success.

\subsection{What Transfers Between Robots?}
A source robot's cup-grasp command specifies its shoulder, elbow and wrist configuration. Native reconstruction anchors its own interface to that command. A foreign decoder receives supervision from the calibrated grasp pose and opening and can select its own native configuration. Installation transforms place both kinematic roots in the task frame, and each FK function carries its chain's link geometry and joint conventions. Tool calibration identifies the point and orientation being compared. The experiments use fixed calibrated installations for each known robot. A different installation would require its corresponding transform in Eq.~\ref{eq:fk}.

The task-frame target is unchanged by camera placement when the demonstrated motion is fixed. The visual policy must still accommodate deployment viewpoints. Camera-frame targets would require the corresponding extrinsics.

Shared tool targets support both learned native decoding and direct pose prediction with online IK. Native reconstruction teaches a decoder to retain configurations represented in its robot's demonstrations. Direct tool prediction delegates joint realization to an online solver. We evaluate task success and command-generation latency for these two routes. The Source/Cross comparison holds the learned-interface route fixed and isolates its foreign geometric supervision.

Foreign realization measures whether source actions can be decoded into the intended tool motion. Visual closed-loop execution then tests whether a policy can use the target robot's decoder on its own observations to complete the missing combination.

\figone{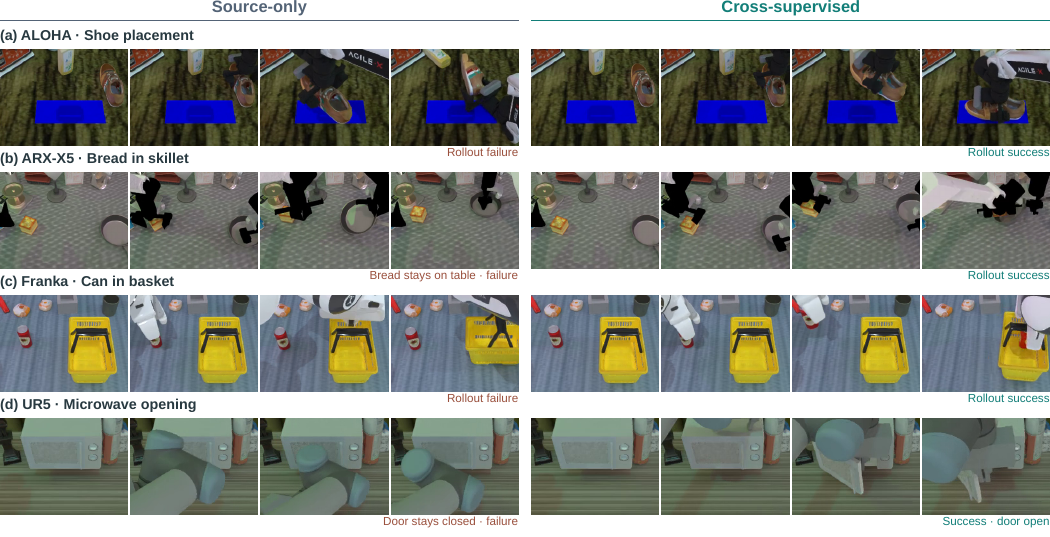}{Selected paired executions with matched initial scenes. Each side shows four recorded stages. Labels report the complete execution outcome.}{fig:current8K-breadth}{}

\subsection{Learning When to Use the Shared Actions}
After interface training, we freeze the encoder and all native decoders. The policy's supervised target is the posterior mean of each demonstrated future action, transformed by a shared TRAIN-fitted diagonal affine normalization $A$,
\begin{equation}
 z_0=A(\mu_s(q^*_{s,1:H})).
\end{equation}
The interface supplies shared action targets and their native realization through frozen robot-specific decoders. It leaves observation-dependent skill prediction to the policy backbone. A predictor trained in these coordinates can execute through the corresponding decoder.

The shared coordinates retain command-level granularity. Temporal dependencies are modeled by the policy's sequence predictor, which learns a horizon of shared actions. At each chunk position, the executing robot's decoder applies the same framewise map to native commands. Interface training can therefore pool command instants across demonstrations with different motion durations.

Our three evaluated backbones instantiate flow learning. Gaussian noise $\epsilon$ and noise level $\sigma$ define $z_\sigma=(1-\sigma)z_0+\sigma\epsilon$. The policy predicts the velocity target $\epsilon-z_0$ with its native weighted flow objective,
\begin{equation}
 \mathcal L_{\mathrm{flow}}=
 w(\sigma)\|v_\theta(z_\sigma,\sigma,c)-(\epsilon-z_0)\|^2,
\end{equation}
where $c$ contains the observation conditioning. At deployment, a predicted chunk is transformed by $A^{-1}$ and decoded through the executing robot's $D_b$. The robot executes a prefix and observes again. Decoded commands specify absolute joint-position targets and gripper opening.

The visual backbone supplies object localization, contact timing and observation-dependent correction. Cross-robot interface training makes source actions meaningful to another robot's native decoder. Our hypothesis is that these two learned components compose on a missing robot--task pairing. Closed-loop task success measures that composition.

\subsection{An Additional Geometric Loss During Policy Learning}
A policy exposing a differentiable clean-action estimate can also receive geometric supervision through the frozen interfaces. For flow predictors, the estimate is recovered from the noisy input and predicted velocity,
\begin{equation}
 \widehat z_0=z_\sigma-\sigma v_\theta(z_\sigma,\sigma,c),\qquad
 \widehat q_b=D_b(A^{-1}(\widehat z_0)).
 \label{eq:clean}
\end{equation}
Applying calibrated FK to $\widehat q_b$ gives predicted tool targets for the source and sampled feasible foreign robots. Their discrepancy from the source demonstration defines $\mathcal L_{\mathrm{FK}}$, using the same pose and opening terms. The action objective becomes $\mathcal L_{\mathrm{flow}}+\lambda_{\mathrm{FK}}\mathcal L_{\mathrm{FK}}$. Frozen decoder and FK parameters retain derivatives with respect to their inputs, so this loss updates the policy through the clean prediction. Targets and feasibility masks are detached and used only during training. The supplementary implementation specifies source-group averaging and valid-target normalization.

The policy-stage loss compares predicted tool poses and openings with demonstration-derived targets through frozen native decoders and calibrated FK. Equal-sized latent deviations can produce different tool displacements because the decoded tool map has direction-dependent sensitivity. This path supplies physical weighting of coordinate-space prediction errors. Its target remains the source demonstration, so interface reconstruction residuals also contribute to the geometric discrepancy. Cross/Full tests this supervision with the learned action correspondence held fixed.

We separate these two opportunities for geometric supervision experimentally. \emph{Source} learns the interface with $\gamma=0$ and trains the policy with $\lambda_{\mathrm{FK}}=0$. \emph{Cross} enables foreign-interface supervision and keeps $\lambda_{\mathrm{FK}}=0$. \emph{Full} uses the same cross-trained interface and enables the policy-stage loss. Source and Cross match action-pretraining data, initialization and exposure. Cross and Full share the frozen interface. These comparisons distinguish learning a transferable action interface from geometrically correcting a policy's clean predictions. Direct EEF matches Cross in visual, language and state conditioning and in its flow objective. It predicts task-frame tool poses and openings, then uses joint-limited numerical IK to obtain native commands. The supplementary implementation gives its representation, solver and fallback settings.

\begin{table*}[t]
\centering
\caption{Cross-robot skill composition across policy backbones. Task-averaged success (\%) on seen and held-out robot--task combinations in RoboTwin and ManiSkill3.}
\label{tab:main}
\small
\setlength{\tabcolsep}{3pt}
\begin{tabular*}{\textwidth}{@{\extracolsep{\fill}}lrrrrrrrr@{}}
\toprule
& \multicolumn{4}{c}{RoboTwin} & \multicolumn{4}{c}{ManiSkill3 (Push)}\\
\cmidrule(lr){2-5}\cmidrule(l){6-9}
& \multicolumn{2}{c}{Native joints} & \multicolumn{2}{c}{Ours (Full)} & \multicolumn{2}{c}{Native joints} & \multicolumn{2}{c}{Ours (Full)}\\
\cmidrule(lr){2-3}\cmidrule(lr){4-5}\cmidrule(lr){6-7}\cmidrule(l){8-9}
Backbone & Seen & Held & Seen & Held & Seen & Held & Seen & Held\\
\midrule
$\pi_{0.5}$ & 97.1 & 17.0 & \textbf{97.4} & \textbf{33.6} & \textbf{82.8} & 3.1 & 79.7 & \textbf{35.9}\\
OpenWAM & 75.3 & 34.8 & \textbf{78.7} & \textbf{46.5} & 80.4 & 11.6 & \textbf{81.3} & \textbf{43.5}\\
FastWAM & 82.9 & 9.7 & \textbf{84.0} & \textbf{29.8} & 78.9 & 5.4 & \textbf{80.1} & \textbf{28.6}\\
\bottomrule
\end{tabular*}
\end{table*}

\newpage
\section{Experiments}
\subsection{Composing Skills Across Policy Backbones}
\figone{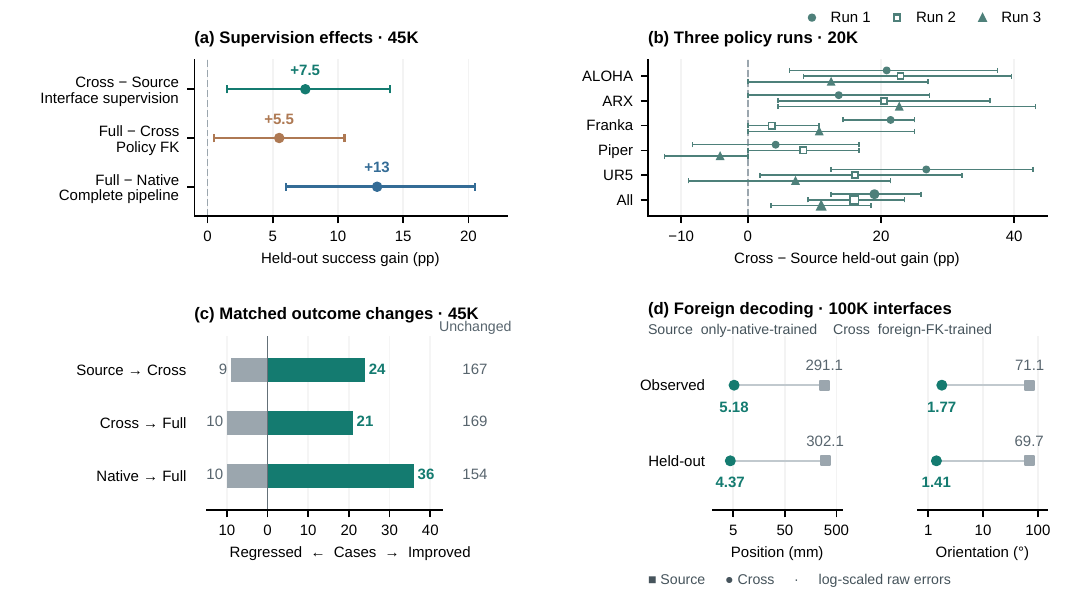}{Independent paired and interface studies. (a) Matched supervision effects. (b) Cross minus Source across three policy runs. (c) Paired scene improvements, regressions and unchanged outcomes. (d) Foreign-decoding errors after native-only (Source) or foreign-FK (Cross) interface training. Intervals in (a,b) resample task clusters.}{fig:analysis}{}

\begin{figure*}[t]\centering
\includegraphics[width=\textwidth]{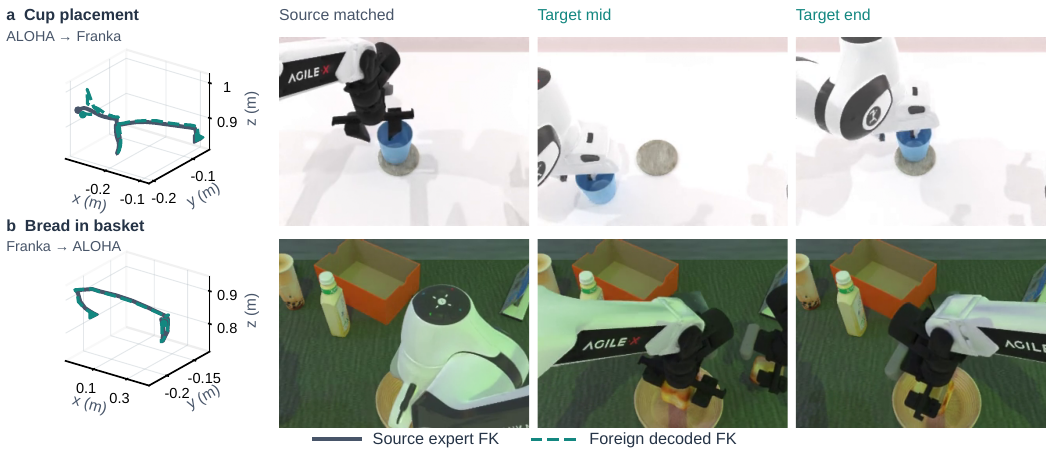}
\caption{Two successful cross-robot expert-command replays through frozen interfaces. Active-arm command FK compares source and foreign tool motion. Source frames match the target command index.}\label{fig:realization}
\end{figure*}
\begin{figure*}[t]\centering
\includegraphics[width=\textwidth]{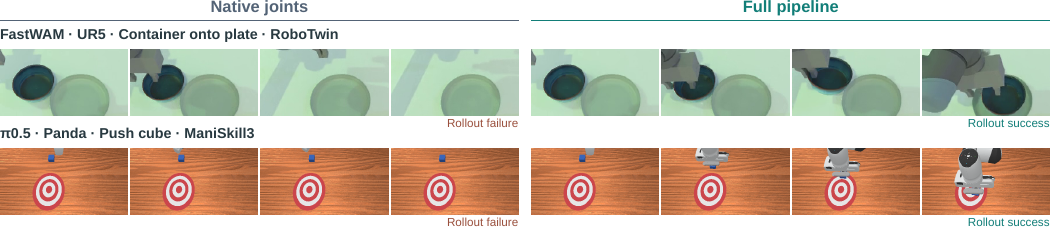}
\caption{Four stages of held-out policy executions with matched initial scenes and crops. Labels report complete execution outcomes. The ManiSkill panel renders recorded simulator states.}\label{fig:cross-backbone-execution}
\end{figure*}

Table~\ref{tab:main} compares native-joint prediction with Full across three policy backbones and two simulation systems. For Tables~\ref{tab:main} and~\ref{tab:mechanism}, we evaluate each task for 1,000 episodes and average task success rates within the seen and held-out groups. The same training framework improves missing-combination transfer with each backbone. Additional controlled studies examine how geometric supervision changes action realization and individual scene outcomes.

Our RoboTwin study~\cite{chen2025robotwin20scalabledata} combines heterogeneous robots and manipulation skills while leaving robot--task edges unobserved. The action interface learns from multi-robot demonstrations that exclude the evaluation combinations. Closed-loop evaluation asks each robot to execute both demonstrated and missing combinations from held-out initial scenes. The supplementary setup gives the demonstration inventories, evaluation panels and model-selection protocol.

Each backbone implements our method through the same learned action interface. A robot's identity selects its native normalization and decoder, and the policy receives images, instructions and current native state. The interface remains frozen during policy learning. Our controlled supervision studies preserve the backbone's flow objective, observation conditioning and training exposure while varying the geometric supervision path. Initialization and implementation details appear in the supplementary setup.

On the RoboTwin $\pi_{0.5}$ and FastWAM groups, Table~\ref{tab:main} shows 82.9--97.1\% seen success for native joints and 84.0--97.4\% for Full, with held-out gains of 16.6--20.1 points. Full also improves seen and held-out success with OpenWAM. Strong performance on demonstrated combinations establishes that the native policies learn the skills. The additional held-out gains show the benefit of connecting those skills to different native realizations.

\begin{table}[t]\centering\small
\caption{Geometric supervision and action realization. Task-averaged success (\%) on seen and held-out robot--task combinations.}\label{tab:mechanism}
\begin{tabular*}{\linewidth}{@{\extracolsep{\fill}}lrrrr@{}}\toprule
& \multicolumn{2}{c}{OpenWAM} & \multicolumn{2}{c}{FastWAM}\\
\cmidrule(lr){2-3}\cmidrule(lr){4-5}
Training & Seen & Held & Seen & Held\\\midrule
Direct EEF + IK & 75.8 & 41.5 & 83.1 & 25.2\\\midrule
Source & 76.3 & 33.4 & 83.9 & 17.7\\
Cross & \textbf{80.1} & 42.9 & \textbf{84.4} & 26.3\\
Full & 78.7 & \textbf{46.5} & 84.0 & \textbf{29.8}\\
\bottomrule
\end{tabular*}
\end{table}

\newpage
Source/Cross fixes interface capacity and demonstration exposure. Cross/Full isolates policy-stage geometric supervision through the same frozen interface.
\newpage
\subsection{How Does Geometric Supervision Help?}
Table~\ref{tab:mechanism} separates action realization from the two stages of geometric supervision. Cross raises held-out success over Source by 9.5 points with OpenWAM and 8.6 points with FastWAM, while retaining strong seen performance. Source already learns a native action interface. The extra correspondence supplied by foreign tool targets makes that interface more useful for missing combinations.

Independent paired studies examine the correspondence at scene level (Figure~\ref{fig:analysis}a--c). Three matched policy runs isolate foreign-interface supervision. Cross improves held-out success over Source by 19, 16 and 11 points, averaging 15.3 points. The positive effect recurs under different policy initializations. Paired outcomes reveal both failures resolved and regressions after adding each supervision stage. These additional trials form a separate panel from the table-level aggregates. The supplementary analysis gives the complete paired outcomes and removes interface-selection pairings while retaining the held-out advantage.

The interface diagnostics isolate geometry before visual control. On certified held-out actions, Source/Cross foreign TCP errors are 302.13/4.37\,mm, while own-body reconstruction remains accurate. Figure~\ref{fig:analysis}d measures foreign realization, Figure~\ref{fig:realization} tests expert-command execution, and Figure~\ref{fig:current8K-breadth} tests visual-policy task completion from each robot's own observations.

Policy-stage FK further raises held-out success by 3.6 points with OpenWAM and 3.5 points with FastWAM (Table~\ref{tab:mechanism}). Cross remains slightly stronger on the seen entries. The clean-prediction loss can correct task-space errors that are weakly penalized in action coordinates, improving the missing combinations in this comparison. Additional learning curves show that policy-stage gains vary across checkpoints. The supplementary results retain the complete fresh-scene and checkpoint comparisons.

\subsection{Native Realization and Further Transfer}
Direct EEF + IK provides a task-space alternative. Full, combining interface and policy-stage supervision, exceeds its held-out success by 5.0 points with OpenWAM and 4.6 points with FastWAM (Table~\ref{tab:mechanism}). Native reconstruction anchors the decoder to demonstrated joint configurations, and native decoding generates commands directly. An independent paired observation-to-command benchmark gives Cross a $1.71\times$ median model-side speedup over Direct EEF + IK. This measurement includes image encoding, flow prediction and native command generation. The supplementary setup gives the timing protocol and solver configuration.

ManiSkill3~\cite{tao2025maniskill3} tests portability in a second simulator. Across the three backbones, Full achieves 79.7--81.3\% seen success and 28.6--43.5\% held-out success, while native-joint policies reach 3.1--11.6\% on the held-out combinations (Table~\ref{tab:main}). This recurring contrast connects the framework's gains to the missing robot--task edges across different policy architectures. Figure~\ref{fig:cross-backbone-execution} illustrates transfer in recorded executions. The supplementary setup gives the task groups and expert-command replays.

\section{Discussion and Conclusion}
Calibrated geometry lets a source demonstration supervise native action realizations across robots. The same relation trains action interfaces and guides policy predictions through their frozen decoders. Matched comparisons across backbones and simulators demonstrate improved transfer to missing robot--task combinations.

\clearpage
\section*{LLM Usage Statement}
OpenAI Codex was used as a supporting tool for code development, analysis, and manuscript preparation. GPT Image was used for the conceptual illustrations in Figure~\ref{fig:teaser}. The authors are responsible for the work.
{\small\bibliographystyle{ieeenat_fullname}\bibliography{references}}
\end{document}